\documentclass[10pt,twocolumn,letterpaper]{article}

\usepackage{iccv}
\usepackage{times}
\usepackage{graphicx}
\usepackage{amsmath}
\usepackage{amssymb}

\usepackage{booktabs}
\usepackage{cuted}
\usepackage{capt-of}

\usepackage{multirow}
\usepackage{booktabs}
\usepackage{tabularx}
\usepackage{caption}
\usepackage{subcaption}
\usepackage{bm}

\usepackage[pagebackref=true,breaklinks=true,letterpaper=true,colorlinks,bookmarks=false]{hyperref}

\iccvfinalcopy 

\begin{document}
	
	\title{Speech Drives Templates:\\Co-Speech Gesture Synthesis with Learned Templates}
	
	\author{
		Shenhan Qian\textsuperscript{\rm 1}\thanks{Equal contribution.} \quad
		Zhi Tu\textsuperscript{\rm 1}\footnotemark[1] \quad 
		Yihao Zhi\textsuperscript{\rm 1}\footnotemark[1] \quad 
		Wen Liu\textsuperscript{\rm 1} \quad 
		Shenghua Gao\textsuperscript{\rm 1,2,3}\thanks{Corresponding author.} \\ \\		
		\textsuperscript{\rm 1}ShanghaiTech University \\
		\textsuperscript{\rm 2}Shanghai Engineering Research Center of Intelligent Vision and Imaging \\
		\textsuperscript{\rm 3}Shanghai Engineering Research Center of Energy Efficient and Custom AI IC
	}

	\maketitle

	\begin{abstract}
	Co-speech gesture generation is to synthesize a gesture sequence that not only looks real but also matches with the input speech audio. Our method generates the movements of a complete upper body, including arms, hands, and the head. Although recent data-driven methods achieve great success, challenges still exist, such as limited variety, poor fidelity, and lack of objective metrics. Motivated by the fact that the speech cannot fully determine the gesture, we design a method that learns a set of gesture template vectors to model the latent conditions, which relieve the ambiguity. For our method, the template vector determines the general appearance of a generated gesture sequence, while the speech audio drives subtle movements of the body, both indispensable for synthesizing a realistic gesture sequence. Due to the intractability of an objective metric for gesture-speech synchronization, we adopt the lip-sync error as a proxy metric to tune and evaluate the synchronization ability of our model. Extensive experiments show the superiority of our method in both objective and subjective evaluations on fidelity and synchronization.
	\footnote{\scriptsize\url{https://github.com/ShenhanQian/SpeechDrivesTemplates}}
\end{abstract}
	\section{Introduction}

\begin{figure}[h!]
	\centering
	\includegraphics[width=1\linewidth]{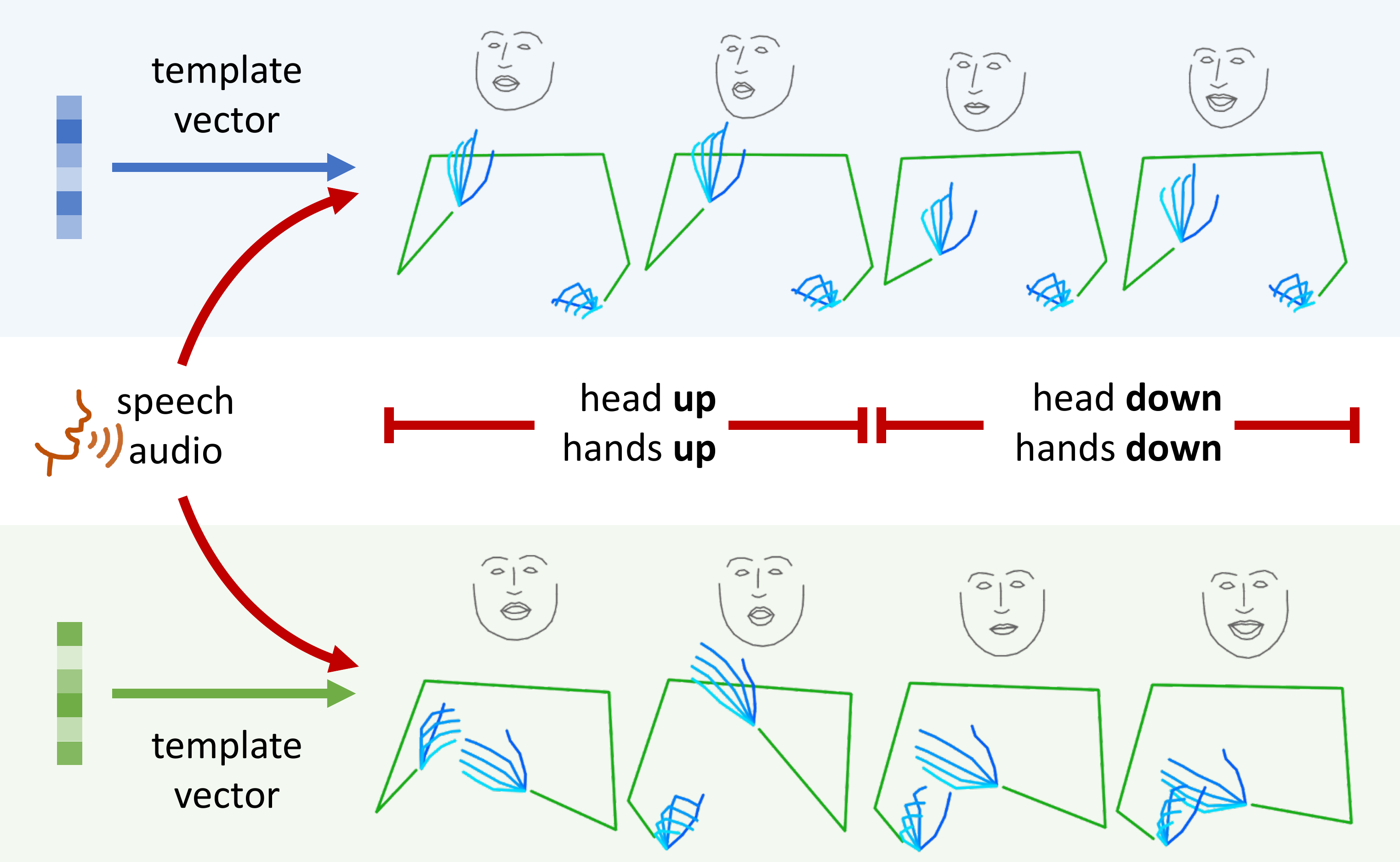}
	\caption{Our method generates realistic gesture sequences from a piece of speech audio. With various template vectors, our method produces two different gesture sequences from the same audio, but the movements are synchronized for the hands, heads, and lips.}
	\label{fig:teaser}
\end{figure}

We humans have always been enthusiastic about making replicas of ourselves. Many successes have been achieved in generation of explicit behaviours, such as lip syncing \cite{suwajanakorn2017synthesizing}, face swapping \cite{thies2016face2face}, or pose re-targeting \cite{chan2019everybody}. But synthesizing implicit behaviors of humans, which plays a key role in synthesizing realistic digital humans, is far less explored.
Co-speech gesture is such a kind of implicit behavior, referring to the movement of body parts when someone is speaking, which conveys rich non-verbal information such as emotion, attitude, and intention. 

Early attempts on co-speech gesture synthesis are mainly rule-based \cite{cassell1994animated, kopp2006towards, WAGNER2014209}, which suffer from poor naturalness because the non-verbal information is too delicate to be described by rules. Later efforts \cite{levine2010gesture, hasegawa2018evaluation, kucherenko2019analyzing, ginosar2019learning, ferstl2019multi, yoon2020speech} go beyond by learning human behaviors from collected data. A non-negligible barrier for data-driven methods is the multi-modal essence of the mapping from speech audio to the possible gestures. This means that for the same input audio, there exist multiple feasible solutions so that directly regressing to the ground-truth gesture casts an inconsistently biased mapping, preventing the model from learning the divergence in the dataset. In recent methods, a common way to cope with this challenge is adversarial learning \cite{ginosar2019learning, ahuja2020style, yoon2020speech} with discriminators narrowing the gap between generated and real ones. However, this can only improve the realism of gestures and have nothing to do with or even harm gesture-speech syncing. Therefore, as long as we expect a stable syncing quality, the regression loss should be the core supervision.

\begin{figure}
	\centering
	\begin{subfigure}[b]{0.48\linewidth}
		\centering
		\includegraphics[width=\linewidth]{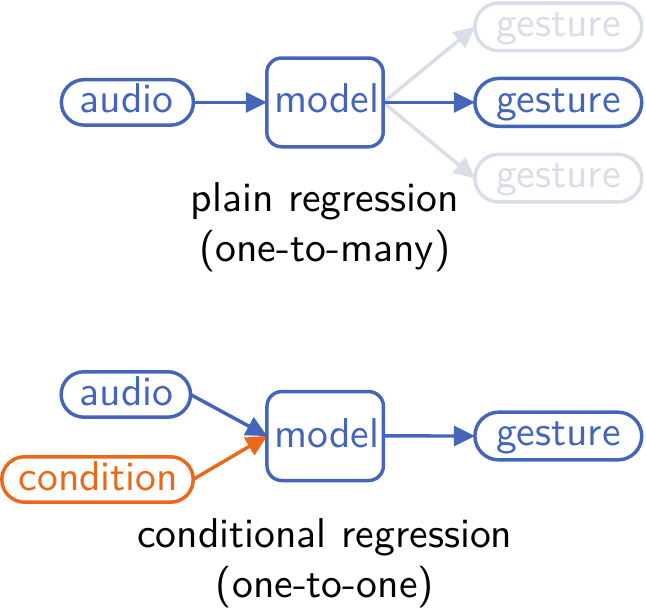}
		\caption{Plain regression versus conditional regression.}
		\label{fig:conditional_reg}
	\end{subfigure}
	\hfill
	\begin{subfigure}[b]{0.46\linewidth}
		\centering
		\includegraphics[width=\linewidth]{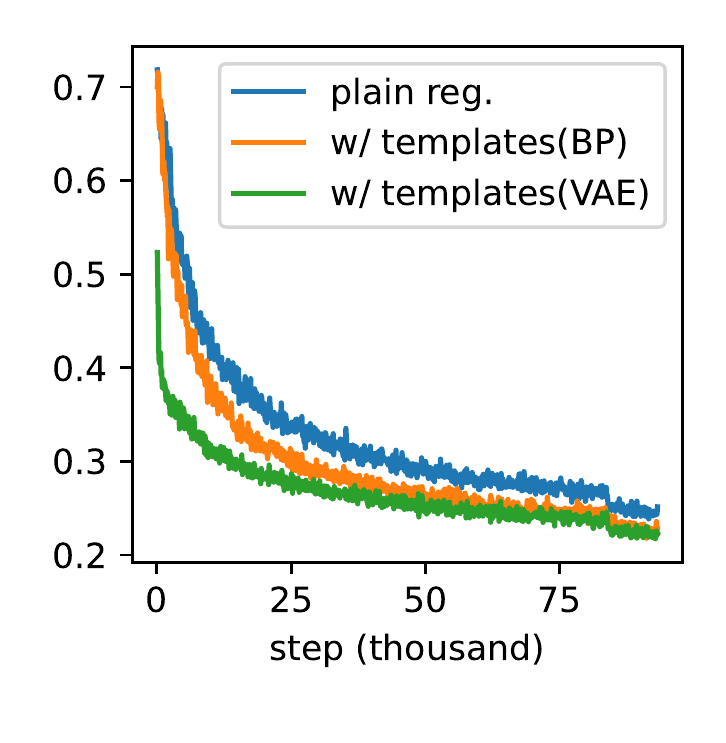}
		\caption{Regression loss curves on the training set.}
		\label{fig:reg_loss_train}
	\end{subfigure}
	\caption{With template vectors serving as the condition, we transform a plain regression with ambiguity into a conditional regression, resulting in a lower regression loss on the training set.}
	\label{fig:intro}
\end{figure}

Given that the regression loss is the only supervision that we can count on to learn gesture-speech synchronization, and the input audio does not provide enough information to determine a gesture sequence exclusively, we complement the input with a condition vector. This condition vector provides the missing information (e.g., habits, emotion, or previous states) to rule out the gestures other than the ground-truth one, thus transforming the mapping from one-to-many into one-to-one (Figure~\ref{fig:conditional_reg}). Concretely, we assign a zero vector to each paired audio-gesture sequence as the initial condition and update the vector along with the network's parameters to minimize the regression loss when training. The intuition here is that if the network can easily regress to the target gesture sequence merely with the audio, the condition vectors will stay the same; otherwise, the condition vector will update to reveal the discrepancy.

From all the learned condition vectors, we can select one and generate a gesture sequence from any audio clip. By switching the condition vector and the speech audio, we observe an interesting phenomenon: the condition vector plays the role of a gesture template. The condition vector determines the general appearance of gestures in a generated sequence, while the audio input adds subtle movements onto the gesture template to bring it to life and match it with the speech. Therefore, we call these condition vectors template vectors for our method. In Figure~\ref{fig:teaser}, we show two gesture sequences generated from the same speech audio with two different template vectors. The resulting gestures are clearly not the same but still match well in the movements of hands, heads, and lips, exhibiting our method's fidelity, variety, and synchronization ability.

Now that we can learn the template vectors with back-propagation, why don't we directly learn them by reconstruction? Therefore, we train a VAE (Variational Auto-Encoder \cite{kingma2014}) to model the distribution of gesture sequences. With this VAE model, we can encode a ground-truth gesture sequence into a template vector and learn the one-to-one mapping from it and the speech audio to the ground-truth gesture sequence. Also, it is possible to decode a template vector to visualize its corresponding gesture sequence. With either back-propagation or VAE, we learn a set of template vectors that not only help lower the regression loss when training (Figure~\ref{fig:reg_loss_train}) but also make generation with variety possible since we can sample one from the learned template vectors to manipulate the general appearance of a synthesized gesture sequence.

Although previous work on co-speech gesture \cite{ginosar2019learning, ahuja2020style, yoon2020speech} limits the scope of gesture to hands and arms only, we advocate including head motion into co-speech gesture, not only for a more unified and harmonized synthesis of the upper body but also for the ease of evaluation. Due to the vagueness of gesture-syncing, existing work heavily rely on subjective evaluations. We propose to adopt the lip regression error as a proxy metric under the hypothesis that to learn gesture syncing well, a model should be able to learn good lip-syncing, as they both depend on the speech, and the latter one is much more deterministic. Furthermore, to assess the fidelity of generated gesture sequences, our trained VAE can be used to compute a Fréchet Template Distance (FTD) similar to the FGD proposed by Yoon \etal \cite{yoon2020speech}, measuring the distribution similarity between the generated ones and the real ones in the feature space.

Our contributions can be summarized as follows:
\begin{itemize}
	\item We propose an audio-driven gesture synthesis method in a conditional learning manner. With the learning of template vectors, we relieve the ambiguity of co-speech gesture synthesis, enhancing the fidelity and variety without sacrificing synchronization quality.
	
	\item We objectify the evaluation of gesture-syncing by borrowing the lip-sync error as a proxy metric. Also, we propose the Fréchet Template Distance (FTD) to assess gesture fidelity.
	
	\item We show the superior synthesis quality of our method in both subjective and objective tests and provide intuitive visualizations of the learned template vectors.
\end{itemize}

	\section{Related Work}
\textbf{Co-Speech Gesture Synthesis.}
Synthesizing co-speech gesture has been an active topic in robotics \cite{10.1145/3308532.3329472, 10.1145/3267851.3267878, yoon2019robots}, graphics \cite{alexanderson2020style, yoon2020speech}, and vision \cite{ginosar2019learning, ahuja2020style, liao2020speech2video}. A recent trend in this task is using in-the-wild videos \cite{ginosar2019learning, ahuja2020style, yoon2019robots} rather than those collected in lab scenarios with sensors, extending the variety of the synthesized gestures. However, as stated by Ginosar \etal \cite{ginosar2019learning}, a barrier on the way towards realistic co-speech gesture generation is the ambiguity of the task, which leads to under-fitting of the data and lack of expressiveness of the results. Although adversarial learning can be incorporated to enhance gesture fidelity as done by Ginosar \etal \cite{ginosar2019learning}, the model still heavily relies on the regression loss to produce gestures synchronized with the audio, so the result is deterministic with no variety. Ahuja \etal \cite{ahuja2020style} disentangle the style and content of gestures by embedding every gesture into a common style space across subjects, achieving style transfer or preserving by switching the style embedding. However, the styles are defined in a per subject manner, featuring only one typical gesture for each subject. 
Alexanderson \etal \cite{alexanderson2020style} introduce the probabilistic model MoGlow based on normalizing flows \cite{papamakarios2019normalizing} to model the mapping from gestures to Gaussian distributions conditioned on the input audio. This model samples latent vectors from Gaussian distributions when inferencing, thus is capable of modeling the one-to-many mapping elegantly. However, the normalizing flows \cite{papamakarios2019normalizing} model only supports linear operations, limiting the expressiveness of the model. Our model relieves the ambiguity of one-to-many mapping with template vector learning and accomplishes diverse generation by sampling the template vector when inferencing.

Besides the method to generate, another huge challenge is evaluation. Due to the ambiguity of co-speech gestures, previous methods mostly rely on the human study to exhibit the effectiveness of their methods \cite{ginosar2019learning, ahuja2020style, alexanderson2020style}, which is reasonable but not objective. As to the objective metrics such as $\mathcal{L}_1/\mathcal{L}_2$ distance, PCK(Percent of Correct Keypoints) reported in \cite{ginosar2019learning, ahuja2020style, alexanderson2020style, yoon2020speech}, they are all based on the distance between the generated and the ground-truth, forming a contradiction between lower error and greater variety. An inspiring attempt on objective metrics is FGD (Fréchet Gesture Distance) proposed by Yoon \etal \cite{yoon2020speech}, which measures the distribution similarity in the feature space. 

\textbf{Talking Head and Lip-Syncing.}
Unlike previous methods on co-speech gesture synthesis, we treat the head as a part of the gesture, not only for the completeness of an upper-body agent but also for the indispensable non-verbal information conveyed by head movements. Representative face manipulation methods \cite{thies2016face2face, wang2018vid2vid, zakharov2019few, burkov2020neural, gafni2021dynamic} are designed in a pose-transfer paradigm, which inherit face landmarks or model parameters \cite{blanz1999morphable} from a target video.
Another stream of methods \cite{karras2017audio, chung2017you, chen2018lip, vougioukas2019realistic, prajwal2020lip, thies2020neural} focus on manipulation of facial expressions or lips, given a piece of speech audio. Karras \etal \cite{karras2017audio} learned a latent code to model emotional states. Prajwal \etal \cite{prajwal2020lip} train a discriminator in an offline manner to enhance lip syncing. These methods achieve plausible synchronization between the lip and the speech, but they cannot or can only manually control the head pose, producing unmatched head motion. Later, Chen \etal \cite{chen2020talking} and Yang \etal \cite{Yang:2020:MakeItTalk} model head movements explicitly in order to lift the naturalness of the talking head, but the synchronization quality is not further evaluated.

	\begin{figure*}[t]
	\centering
	\includegraphics[width=1\linewidth]{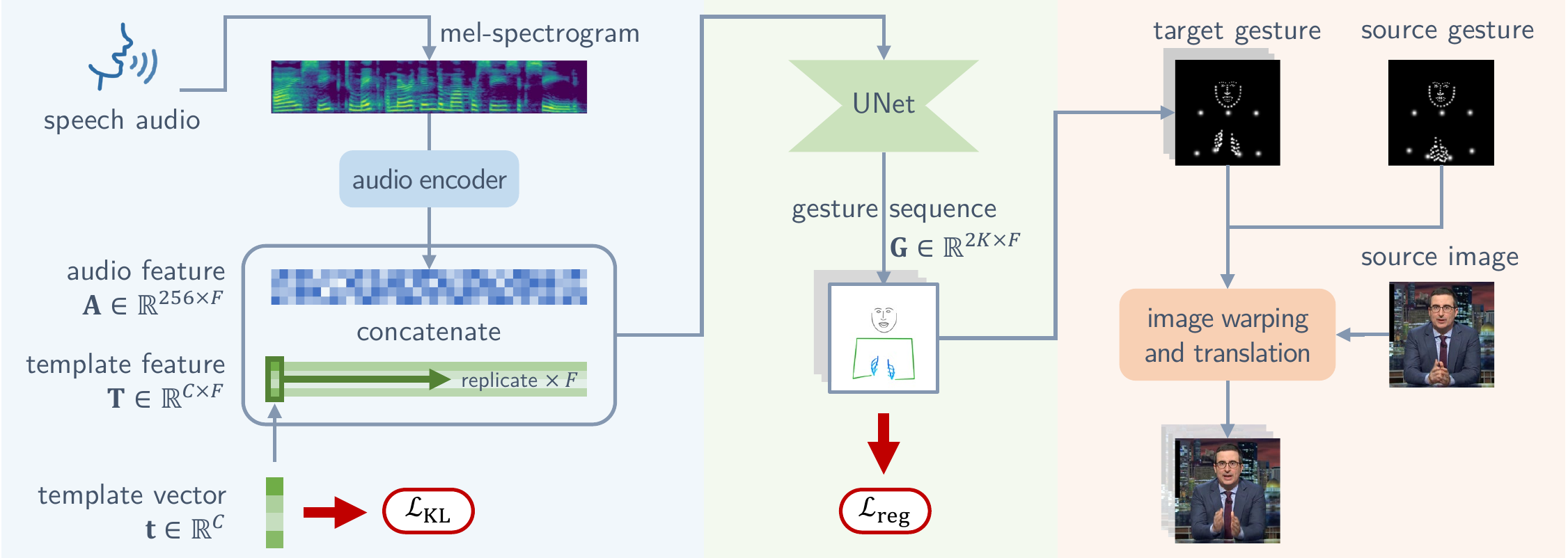}
	\caption{Our network takes an audio feature $\mathbf{A}$ and a template feature $\mathbf{T}$ as the input to generate a co-speech gesture sequence $\mathbf{G}$. The template vectors $\mathbf{t}$ are updated with back-propagation or encoded by a VAE. $F$ is the number of frames, $C$ is the dimension of the template vector, and $K$ is the number of keypoints. With the generated gesture sequence as an intermediate representation, we can synthesize a realistic video with an image warping and translation module.}
	\label{fig:arch}
\end{figure*}

\section{Methods}
\textbf{Pipeline.}
Given a piece of speech audio as the input, we generate a sequence of gestures with natural poses and synchronized motions. An overview of our model is shown in Figure~\ref{fig:arch}. 

Formally, for an audio clip, we follow previous methods \cite{ginosar2019learning, alexanderson2020style} to transform the audio waveform into a mel-spectrogram, which is a 2D map with time and frequency on each axis. Then we send it into an audio encoder to get the audio feature $\mathbf{A} \in \mathbb{R}^{256 \times F }$, where $F$ is the number of frames. 
As another input, our network takes in a template vector $\mathbf{t} \in \mathbb{R}^{C}$ and stacks the replicas of it into a template feature $\mathbf{T} \in \mathbb{R}^{C \times F}$ to align the timeline of the audio feature. Therefore, the complete input of our model is $\left[\mathbf{A|T}\right]\in \mathbb{R}^{(256+C)\times F}$, the concatenation of the audio feature $\mathbf{A}$ and the template feature $\mathbf{T}$.

Our gesture generation network is a UNet-alike 1D convolutional neural network that slides along the timeline with a 7-layer encoder, a 6-layer decoder, and skip-connections. 

The output is a gesture sequence $\mathbf{G} \in \mathbb{R}^{2K \times F}$, where $2K$ corresponds to the 2D coordinates of $K$ upper-body keypoints in a frame, including the face, hands, and arms.

As a main supervision, we apply $L_1$ regression loss on a regressed gesture sequence $\mathbf{G}$:
$$ \mathcal{L}_{\text{reg}} = \frac{1}{F} \sum_{i=1}^{F}{\| \mathbf{G}^{(i)} - \hat{\mathbf{G}}^{(i)} \|_1}, $$
where $\mathbf{G}^{(i)}$ and $\hat{\mathbf{G}}^{(i)}$ are the predicted and ground-truth gesture vectors in the $i^{th}$ frame of $\mathbf{G}$.

\textbf{Image Synthesis.}
To ease visual evaluation of our generated gesture sequences, we train an image warping and translation module, inspired by Balakrishnan \etal \cite{balakrishnan2018synthesizing}. For each frame, we first warp pixels of each body part in the source image to the target location with local affine transformations to obtain a coarse result, then feed the concatenation of the coarse result and heatmaps of keypoints into an image translation network as a refinement. During the train session, source and target pairs are randomly combined. 

\subsection{Complement Audio with Learned Conditions}
\label{sec:intro_tempalte_vector}
We mainly rely on the regression loss to train our model because it is the only reliable source of supervision towards gesture sequences synchronized with the audio. However, since the mapping from the speech audio to a gesture sequence is not exclusive, \ie, there exist many other feasible gestures, simply regressing to the ground-truth gesture sequence causes ambiguity and leads to overly smooth results.

To remove the ambiguity, we should provide more information to our model. Concretely, we additionally feed in a condition vector, as shown in Figure~\ref{fig:conditional_reg}. Here, we expect the condition vector to narrow the range of potential gestures instead of pointing to a specific static gesture; otherwise, the role of the input audio will be weakened, which harms gesture-speech syncing. To this end, we assign one condition vector to each short gesture sequence (about 4 seconds long) instead of each frame and regress from the audio and the condition vector to the ground-truth gesture sequence. This relieves the ambiguity between speech and gestures, laying the foundation of our method.

We call this condition vector a template vector for our method because this vector determines the general appearance of the generated gesture sequence, while the input audio adds subtle movements to match the speech and the gesture sequence, just like the relationship between the template and the content.

\textbf{Learning Templates by Back-Propagation.} 
We initialize the template vector of each speech-gesture pair to a zero vector, supposing them to be subject to the same condition. When training, we back-propagate the regression loss and update template vectors along with parameters of the UNet. This means that the model is trained without extra information since all the template vectors are set to zero; when ambiguity occurs, the template vector will be updated to relieve the ambiguity. By storing the trained template vectors, we extract the latent condition for each sample from the dataset.

To regularize the template vector space, we apply a KL-divergence loss
$$\mathcal{L}_\text{KL} = D_\text{KL}\left( \mathcal{N}\left(\mu_\mathbf{t}, \sigma^2_\mathbf{t}\right) \|  \mathcal{N}\left(0, 1\right) \right)
$$
where $\mu_\mathbf{t} \in \mathbb{R}^{C}$ and $\sigma^2_\mathbf{t} \in \mathbb{R}^{C}$ are the mean and variance vectors of template vectors $\mathbf{t}$ in a mini-batch. Then the total loss function is defined as follows:
\begin{equation*}
	\mathcal{L} = \lambda_{\text{reg}}\mathcal{L}_{\text{reg}} + 	\lambda_{\text{KL}}\mathcal{L}_{\text{KL}},
\end{equation*}
where $\lambda_{\text{reg}}$ and $\lambda_{\text{KL}}$ are weights applied to the loss terms. We set $\lambda_{\text{KL}}=1$ and $\lambda_{\text{reg}}=1$ in our experiments.

Updating template vectors by back-propagation brings several benefits. First, the regression loss converges faster lower than a plain regression from the audio, indicating a better fitting of the training set (see Figure~\ref{fig:reg_loss_train}). 
Second, our model can generate diverse gestures by sampling arbitrary template vectors from the trained ones while maintaining highly synchronized gestures and lips. 
Third, interpolation of template vectors results in smooth changes of gestures, such as switching hands and changing the head orientation, demonstrating a compact condition space.

Despite the above benefits and inspirations, this method still has some limitations. 
First, since the template vector is assigned in a sample-wise way, each template will only be used and updated once an epoch, which requires careful tuning of hyper-parameters (e.g., learning rate, number of epochs) to let the template vectors converge. 
Second, although we can observe gesture variations caused by template switching, we still lack interpretation of the templates. 
Third, we can only conduct the mapping from a template vector to a gesture sequence but cannot go in the opposite direction.

\textbf{Learning Templates by Reconstruction.}
\label{sec:sdt_vae}
To resolve the above limitations, we consider learning template vectors by reconstruction with VAE \cite{kingma2014}. This VAE first encodes a ground-truth gesture sequence $\hat{\mathbf{G}}$ into a mean vector $\mu_\mathbf{t} \in \mathbb{R}^{C}$ and a variance vector $\sigma^2_\mathbf{t} \in \mathbb{R}^{C}$, then decode them into a reconstructed gesture sequence $\mathbf{G}$. Similarly, it is built up with fully 1D convolutions sliding along the timeline and is also trained with a $L_1$ loss and a KL-divergence loss. 

Once the VAE is trained, it is frozen to be used as a template vector extractor for computation of FTD described in Section \ref{sec:evaluation}.

\subsection{Evaluation of Co-Speech Gesture Generation}
\label{sec:evaluation}
Common evaluation metrics used by prior methods for co-speech gesture like $L_1/L_2$ distance, accuracy, or PCK (percent of correct keypoints) are not ideal because what they measure is the distance between a generated sample and the ground truth, ignoring the diversity of feasible gestures for a given audio clip. Therefore, targeting these metrics will result in boring and inexpressive synthesis. 

Intuitively, a good gesture synthesis should meet at least two requirements: naturalness and synchronization; but none of them can be easily measured with distance-based metrics. Next, we propose two metrics for gesture assessment in terms of synchronization and naturalness, respectively.

\textbf{Lip-Sync as A Proxy Metric.}
Different from body gestures that are diverse, the lip shapes are almost consistent because pronouncing a syllable usually requires a particular mouth shape. Also, we observe better convergence of lip keypoints' regression losses than others on the validation set, which confirms the consistency of the mapping from speech audio to the lip. 

Therefore, we adopt the distance between generated lip keypoints and the ground-truth as a proxy metric for synchronization measurement of the entire gesture. This proxy metric works for two reasons: 1) Both lip keypoints and other keypoints share the same network and features, and our method has no special design for lip syncing; 2) Although good lip syncing quality cannot guarantee good gesture quality, but lip syncing degradation is a good warning signal of bad gesture-syncing.

Formally, the proxy metric we use is the normalized lip-sync error
\begin{equation}
	\mathcal{E}_{\text{lip}} = \frac{\frac{1}{F} \sum_{i=1}^{F} \| d^{(i)} - \hat{d}^{(i)} \|_2}{\max_{1 \leq n \leq F} \hat{d}^{(n)}}
\end{equation}
where $d^{(i)}$ is the distance between the center keypoints of upper and lower lip in the $i^{th}$ frame of the generated gesture sequence $\mathbf{G}$, and $\hat{d}^{(i)}$ is the corresponding distance for ground-truth gesture sequence $\hat{\mathbf{G}}$.

\textbf{Fréchet Template Distance.}
As aforementioned, directly measuring the distance between a generated gesture sequence and the ground truth discourages variety. Here, we introduce FTD (Fréchet Template Distance) as a variation of FID (Fréchet Inception Distance) \cite{heusel2017gans}. FTD measures the distribution distance between the synthesized and the real gesture sequences among a group of samples rather than a single sample. Therefore, in order to achieve a better FTD score, the generated results should be not only natural but also diverse.

In our experiments, FTD is computed on the entire test test as follows:
\begin{equation*}
	\text{FTD}=\left| \mu_{\mathbf{t}} - \mu_{\hat{\mathbf{t}}} \right|^2 + \text{tr}\left( \Sigma_{\mathbf{t}} + \Sigma_{\hat{\mathbf{t}}} - 2\left( \Sigma_{\mathbf{t}}  \Sigma_{\hat{\mathbf{t}}} \right)^{1/2} \right),
\end{equation*}
where $\mu_{\mathbf{t}}$ and $\Sigma_{\mathbf{t}}$ are the mean vector and covariance matrix of the template vectors $[ \mathbf{t}_1, \mathbf{t}_2, \dots, \mathbf{t}_N ]$, encoded from synthesized gesture sequences $[ \mathbf{G}_1, \mathbf{G}_2, \dots, \mathbf{G}_N ]$ across the test set with the VAE described in Section~\ref{sec:sdt_vae}, where $N$ denotes the number of samples in the test set. $\mu_{\hat{\mathbf{t}}}$ and $\Sigma_{\mathbf{\hat{\mathbf{t}}}}$ are the counterparts for the ground truth.

	\section{Experiments}

\textbf{Datasets.}
We test our method on the Speech2Gesture \cite{ginosar2019learning} dataset since it is the only one providing complete annotations for the upper body, especially for face keypoints. 
However, we only report results of two speakers, Oliver and Kubinec, of this dataset due to the unusable quality of the face and hand keypoints of other speakers (pseudo labels acquired with OpenPose \cite{cao2019openpose}). For results on other speakers, please refer to the supplementary material. 
Furthermore, we collect data of two Mandarin speakers, Xing and Luo, to test the versatility of our method.
The video clips of the four speakers have a length of about 25.13 hours in total after manual filtering of incorrect annotations. We train our model on each of the speakers separately because we focus on speaker-specific gesture learning.

\textbf{Evaluation.}
We report three objective metrics for fair comparison: 1) the $L_2$ distance that directly measures the distance between the prediction and the ground truth; 2) the normalized lip-sync error ($\mathcal{E}_{\text{lip}}$) as a proxy metric for gesture synchronization; 3) the Fréchet Template Distance (FTD) as a measurement of fidelity. 

We conduct an extensive human study to perceptually compare our method with baselines and verify the feasibility of our proposed objective metrics. 
We create videos with gesture sequences generated by different methods from the same pieces of speech audio, then publish them as an online questionnaire for human evaluation. For each of the four speakers, we randomly sample 8 pieces of speech audio for video generation. For each questionnaire, we randomly choose at least 2 of the 8 videos for each speaker to form a questionnaire with 10 video clips.
During a test, a participant is exposed to the 10 videos one by one. Each video shows the results of competing methods synchronously with audio. The results are anonymized by letters and visualized with both skeleton maps and synthesized images. After watching each video, a participant is asked to make three choices: 1) the one with the best lip-sync quality; 2) the one with the best gesture-sync quality; 3) the one with the most natural gestures. The final result is computed by the averaged percentage of each method selected as the best in each question. After the test, we collect 65 effective questionnaires in total.

\textbf{Implementation Details.}
When preparing data, we segment videos into short clips of 64 frames in 15 FPS (about 4 seconds). 
To eliminate the scale difference across speakers and video resolutions, we re-scale the skeletons of each speaker according to their averaged shoulder widths. 
We set the dimension of template vector space $C=32$ in all the experiments. Although our method is not sensitive to $C$, too large a dimension leads to degradation of gesture-syncing, and too low a dimension limits the expressiveness of the template space.
We use a batch size of 32 for both training and test. We train our model with an Adam optimizer for 100 epochs. We use learning rate 0.0001, and downscale it for 10 times at the 90$^\text{th}$ and 98$^\text{th}$ epoch.
When testing, we randomly sample a template vector from the trained ones corresponding to clips in the training set, making our results diverse and non-deterministic.

\subsection{Regression with Learned Templates}
\label{sec:regression_with_learned_conditions}

\begin{table}[b]
	\centering
	\caption{Effects of template learning in different settings. Red digits refer to the worst among listed models. Our models with clip-wise templates achieve the best balance between synchronization and expressiveness.}
	\label{tab:regression_with_templates}
	\small
	\begin{tabular}{lccc}
		\toprule
		&template type	&$\mathcal{E}_{\text{lip}} \downarrow$	&FTD $\downarrow$  \\
		\midrule
		w/o template	&-     			&\textbf{0.17}						&\textcolor{red}{1.66} \\
		w/ template-BP	&frame-wise		&\textcolor{red}{0.21}		&0.78 \\
		\midrule
		w/ template-BP	&clip-wise		&\textbf{0.17}				&1.26 \\
		w/ template-VAE	&clip-wise		&\textbf{0.17}					&\textbf{0.92} \\
		\bottomrule
	\end{tabular}
\end{table}

As the core of our method, template vector learning makes it possible to learn the one-to-many mapping from a piece of speech audio to feasible gesture sequences merely with a regression loss. In Table~\ref{tab:regression_with_templates}, we show quantitative comparison between different configurations of templates. The model without templates gets the worst FTD, indicating poor expressiveness of learned gestures. On the contrary, the model with frame-wise template vectors gets the worst lip-sync error ($\mathcal{E}_\text{lip}$), indicating degradation in gesture-syncing. This results from the excessive expressiveness of per-frame template vectors since the model can simply store per-frame gestures in the per-frame template vectors without extracting information from the audio signal. Meanwhile, our models with clip-wise template vector (learned either by back-propagation or VAE) achieve the best balance between synchronization and expressiveness with relatively low lip-sync error and FTD. In other words, our models with clip-wise templates produce more diverse gestures with almost no harm to synchronization. 

\begin{figure}[t]
	\centering
	\includegraphics[width=1\linewidth]{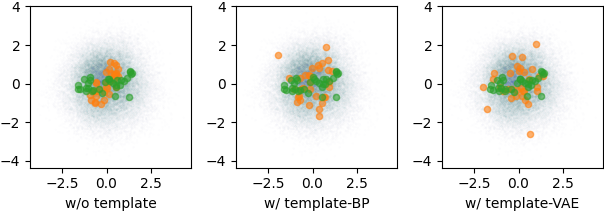}
	\caption{Visualization of ground-truth gestures versus generated gestures in the template space. By feeding in a gesture sequence to the encoder of our trained VAE, we obtain the template vector of it. For visualization, we project the template vectors onto a 2D plane with PCA.}
	\label{fig:batch_tmplt_projection}
\end{figure}

To confirm the variety of our results, we use the encoder of our trained VAE to obtain the corresponding template vectors of both the ground-truth and the generated gesture sequences and visualize them by projection onto a 2D plane via PCA.  As shown in Figure~\ref{fig:batch_tmplt_projection}, for the model without templates, the encoded vectors gather around the origin. In contrast, the encoded vectors of our results from clip-wise templates span a larger space, showing a larger variety, which is consistent with the lower FTD values in Table~\ref{tab:regression_with_templates}.

\subsection{Comparison with Baselines}

\begin{table*}[t]
	\centering
	\caption{Comparison with baselines for gesture generation on two English speakers from the Speech2Gesture \cite{ginosar2019learning} dataset (Oliver and Kubinec) and two Mandarin speakers that we collect (Xing and Luo). Our models produce results with superior synchronization and expressiveness. Note that lower $L_2$ distance does not indicate better performance for our task.}
	\label{tab:related-work}
	\renewcommand{\tabcolsep}{1.6mm}
	\small
	\begin{tabular}{l|ccc|ccc|ccc|rcc}
		\toprule
		&\multicolumn{3}{c|}{Oliver} &\multicolumn{3}{c|}{Kubinec} &\multicolumn{3}{c|}{Xing} &\multicolumn{3}{c}{Luo} \\ \cline{2-13}
		&\multicolumn{1}{c}{$L_2$ dist.} &\multicolumn{1}{c}{$\mathcal{E}_{\text{lip}} \downarrow$} &\multicolumn{1}{c|}{FTD $\downarrow$} &\multicolumn{1}{c}{$L_2$ dist.} &\multicolumn{1}{c}{$\mathcal{E}_{\text{lip}} \downarrow$} &\multicolumn{1}{c|}{FTD $\downarrow$} &\multicolumn{1}{c}{$L_2$ dist.} &\multicolumn{1}{c}{$\mathcal{E}_{\text{lip}} \downarrow$} &\multicolumn{1}{c|}{FTD $\downarrow$} &\multicolumn{1}{c}{$L_2$ dist.} &\multicolumn{1}{c}{$\mathcal{E}_{\text{lip}} \downarrow$} &\multicolumn{1}{c}{FTD $\downarrow$} \\ 
		\midrule
		Audio to Body \cite{DBLP:conf/cvpr/ShlizermanDSK18} 
		&49.7	&0.19			&3.48			&70.9	&0.17			&4.51 				&50.9	&0.18			&4.75		&48.4	&\textbf{0.16}	&2.70\\
		Speech2Gesture \cite{ginosar2019learning} 
		&53.5	&0.23			&8.30	    &64.9	&0.20			&4.53				&48.0	&0.19			&4.49			&63.7	&0.20			&3.10	 \\
		MoGlow \cite{alexanderson2020style} 
		&50.6	&0.20			&2.28	&78.1	&0.16			&2.49 				&48.4	&0.18			&4.94				&54.8	&0.18			&1.47\\
		Ours (w/ template-BP) 
		&50.6	&\textbf{0.17}	&1.26	&83.7	&\textbf{0.15}	&1.98 				&50.0 	&\textbf{0.17}	&2.72				&51.5	&\textbf{0.16} 	&1.21\\
		Ours (w/ template-VAE) 
		&62.4	&\textbf{0.17}	&\textbf{0.92} &100.7	&\textbf{0.15}	&\textbf{1.07}		&57.8	&0.18			&\textbf{1.72}	 &80.8	&0.17			&\textbf{0.69}\\ 
		\bottomrule
	\end{tabular}
\end{table*}

\textbf{Baselines.}
\textit{Speech2Gesture} \cite{ginosar2019learning} is a fully convolutional model that directly regresses from a mel-spectrogram to gesture sequences. To add keypoints of the face, we enlarge the number of channels for the last convolution layer. For the best balance between the regression loss and the adversarial loss, we set the weight for the latter one to 0.1.
\textit{Audio to Body Dynamics} \cite{DBLP:conf/cvpr/ShlizermanDSK18} is a sequential model with separate LSTM \cite{10.1162/neco.1997.9.8.1735} models for regression of body and hand keypoints. We add one more LSTM model for face keypoints. Following the original configuration, we feed in a 28-channel MFCC. We adjust the hidden layer dimension of LSTM models from 200 in the original implementation to 800 for the best performance.
\textit{MoGlow} \cite{alexanderson2020style} is a probabilistic gesture generator based on normalising flows \cite{papamakarios2019normalizing}. We modify its output channels to adapt to our task. For better performance, we feed in mel-spectrogram instead of MFCC, and set the hidden layer dimension $H = 800$ and the number of normalizing steps $K = 12$. 

\label{sec:comparisons}
\textbf{Objective Comparison.}
We compare our models with the above baselines across four speakers. As shown in Table~\ref{tab:related-work}, our models produce the smallest normalized lip-sync error and the smallest FTD on all speakers, indicating superior gesture-syncing and expressiveness. Meanwhile, our models produce relatively high $L_{2}$ distances. This is expected since our results are generated with randomly sampled template vectors, which should not always conform to the ground truth gesture.

\begin{figure}[h]
	\centering
	\includegraphics[width=\linewidth]{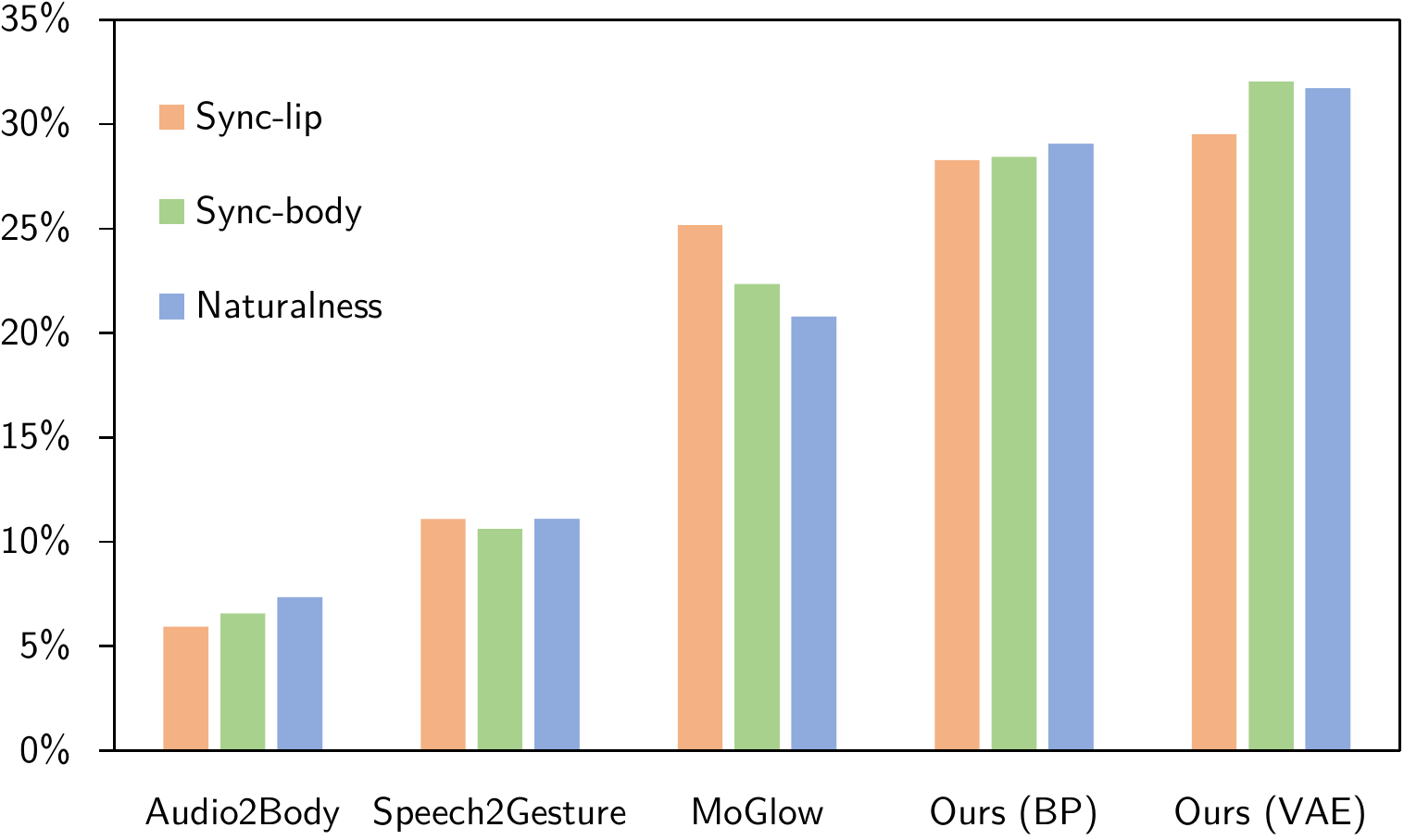}
	\caption{Human study on participants' preference in lip and body synchronization and naturalness among methods (in percentage).}
	\label{fig:user_study}
\end{figure}

\begin{figure*}[t]
	\centering
	\includegraphics[width=1\linewidth]{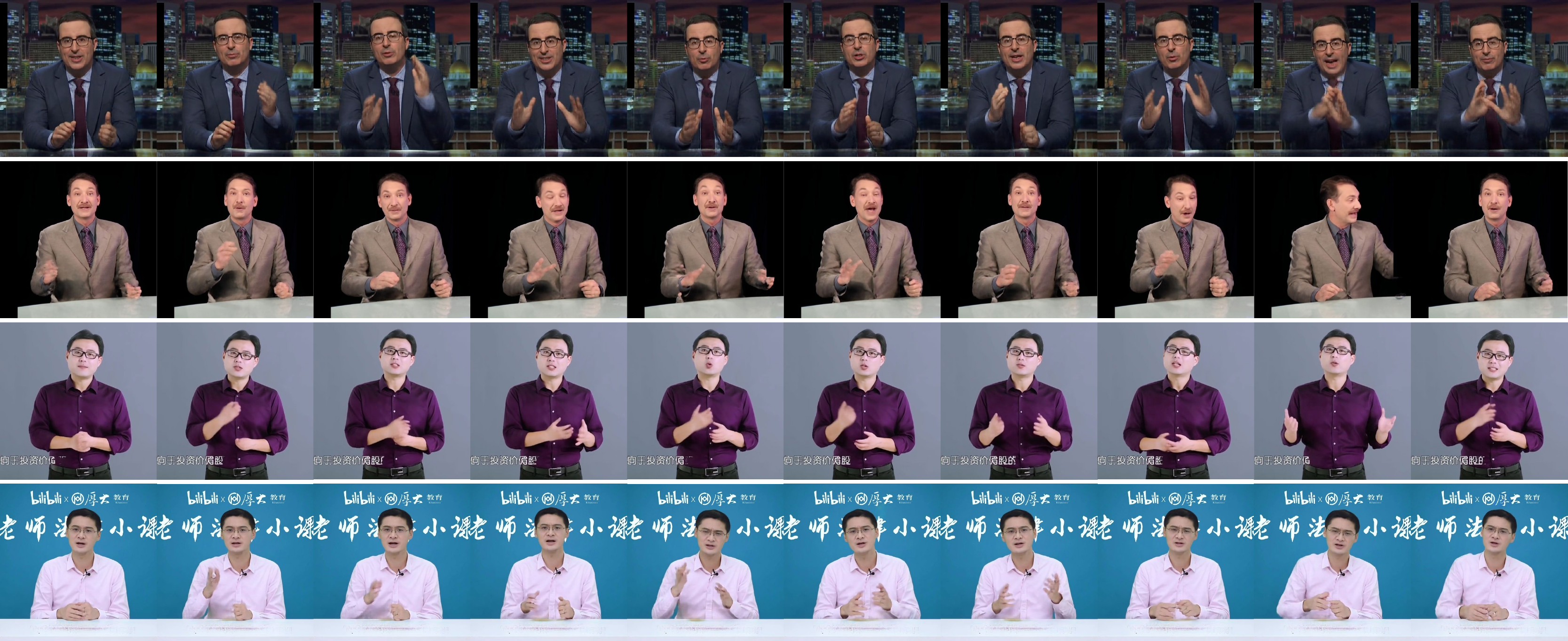}
	\caption{Examples of image frames synthesized from generated gesture sequences with our image synthesis module. In the 1$^\text{st}$ and the 2$^\text{nd}$ lines are Oliver and Kubinec from the Speech2Gestrure \cite{ginosar2019learning} dataset. In the 3$^\text{rd}$ and the 4$^\text{th}$ lines are Xing and Luo that we collect from the Internet.}
	\label{fig:pose2img}
\end{figure*}

\textbf{Subjective Comparison.}
For a perceptual comparison between methods, we invite volunteers to watch anonymized results and choose the best in three aspects. Examples of synthesized images used in the human study are shown in Figure~\ref{fig:pose2img}. According to the bar chart in Figure~\ref{fig:user_study}, our models show significant advantages over baseline models. It is to be mentioned that this human study verifies the strong correlation between the performance of lip syncing and body syncing, which endorses our proposal of adopting the normalized lip-sync error ($\mathcal{E}_{\text{lip}}$) as a proxy metric to measure the extent of how a gesture is synchronous with the speech audio.

\subsection{Visualization of Template Space}
\begin{figure}[t]
	\centering
	\includegraphics[width=1\linewidth]{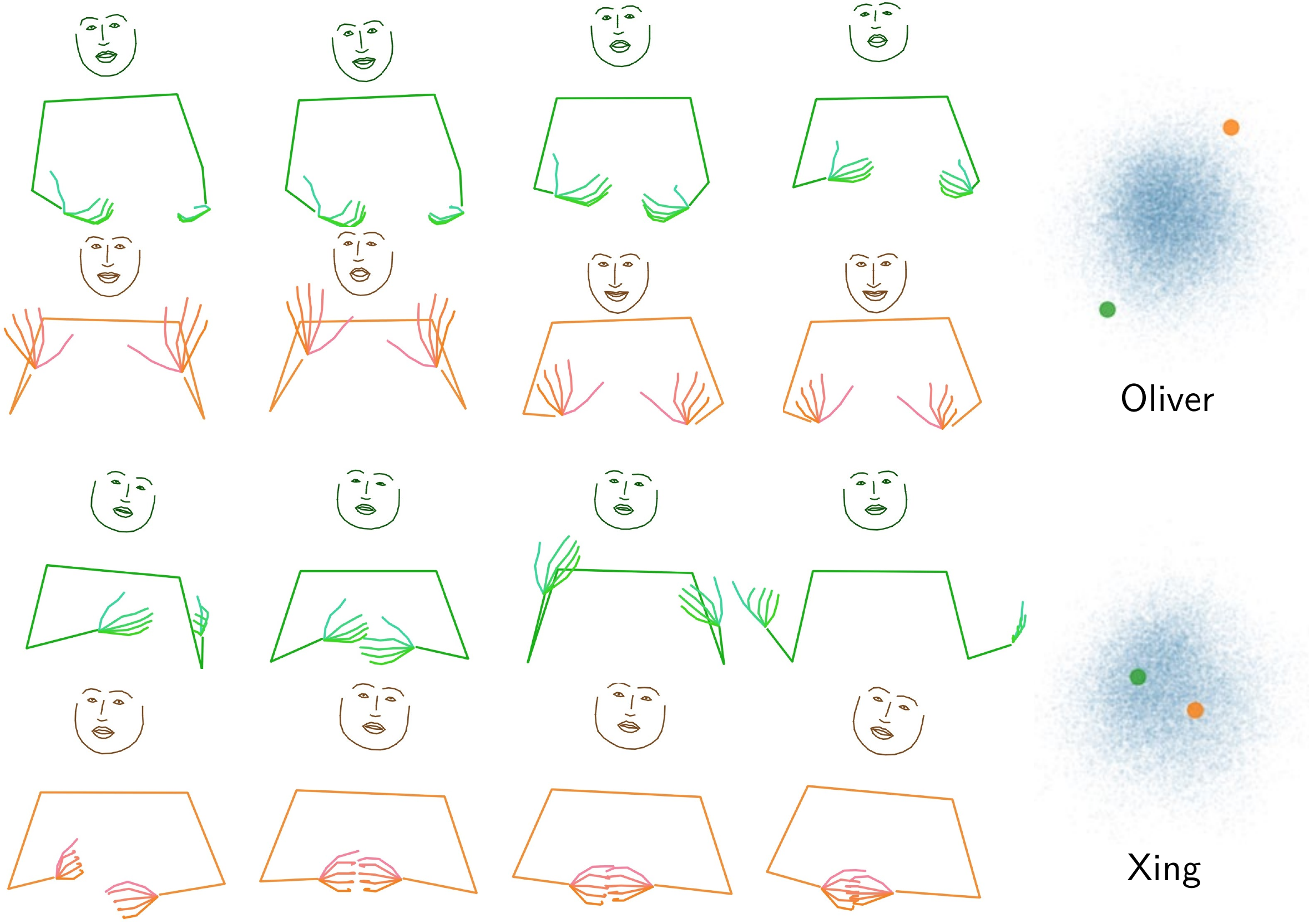}
	\caption{Visualization of the template vector space. Each scatter plot is a projection of the template vector space of a subject, which is close to a Gaussian distribution. The green point in a scatter plot is the endpoint of a sampled template vector, while the orange one refers to its opposite vector. Each line of skeleton sequences corresponds to a template vector by color. For each subject, the gesture sequences of opposite template vectors exhibit clear semantic symmetry.
	}
	\label{fig:gesture_vis}
\end{figure}

For a better interpretation of the template vectors, we explore the property of the vector space. We visualize the corresponding gestures of a particular template vector and its opposite by feeding the vectors into our trained VAE's decoder, respectively. Instead of manually selecting template vectors, we adopt a closed-form factorization algorithm proposed by Shen and Zhou \cite{shen2020closed} for latent semantics discovery. Taking the weight matrix of the first layer of the VAE's decoder, we conduct eigenvalue decomposition and keep the eigenvector with the largest eigenvalue. From the results for Oliver and Xing in Figure~\ref{fig:gesture_vis}, we observe high semantic symmetry such as the position and orientation of heads and hands.

\subsection{Ablation Studies}

\textbf{Transposed Instance Normalization.}
In our experiments, we observe significant improvements on $\mathcal{E}_{\text{lip}}$ and FTD by substituting BN (Batch Normalization) with IN (Instance Normalization) as shown in the $1^\text{st}$ and the $2^\text{nd}$ row in Table~\ref{tab:ablation}. However, models with IN produce results with high-frequent vibrations.
Therefore, we propose the transposed instance normalization (IN$^*$), which conducts normalization on the dimension of keypoints ($B, \bm{2K}, F$) rather than frames ($B, 2K, \bm{F}$). This operation produces stable gesture sequences with comparable performance (the $3^\text{rd}$ row in Table~\ref{tab:ablation}).

\textbf{Hierarchical Gesture Representation.}
Considering the kinematics of the human body, we try to decouple body parts by a hierarchical body representation with separate root nodes for keypoints of the face, arms, and hands. Comparing the $3^\text{rd}$ row and the $4^\text{th}$ row of Table~\ref{tab:ablation}, we can see an obvious improvement on lip syncing ($\mathcal{E}_{\text{lip}}$).

\begin{table}[t]
	\centering
	\caption{Ablation study on the normalization operation and body representation. Experiments are conducted on our model without template vectors on Oliver. 
		IN$^*$ denotes our proposed transposed instance normalization. Hierarchical denotes that gesture keypoints are split into four parts, and each part has its local root node.
	}
	\label{tab:ablation}
	\small
	\begin{tabular}{lccc}
		\toprule
		&Hierarchical	&$\mathcal{E}_{\text{lip}} \downarrow$	&FTD $\downarrow$ \\ 
		\midrule
		BN		&				&0.20									&7.01 \\
		IN		&				&0.19									&1.54 \\
		IN$^*$	&				&0.19									&\textbf{1.53} \\
		IN$^*$ &$\checkmark$	&\textbf{0.17}							&1.66 \\ 
		\bottomrule
	\end{tabular}
	\vspace{-.3cm}
\end{table}
	\section{Conclusion}
This paper aims to synthesize a gesture sequence of the complete upper body given speech audio as input. Based on the fact that speech cannot fully determine gestures, we propose to learn a set of gesture templates, which relieve the ambiguity and increase the variety and fidelity of synthesized gestures. Furthermore, we propose to use the normalized lip-sync error as a proxy metric for gesture synchronization and FTD as a measurement of fidelity. Quantitative and qualitative results on four speakers across two languages show the superiority of our method.

	\section*{Acknowledgments}
	The work was supported by National Key R\&D Program of China (2018AAA0100704), NSFC \#61932020, Science and Technology Commission of Shanghai Municipality (Grant No. 20ZR1436000), and “Shuguang Program” supported by Shanghai Education Development Foundation and Shanghai Municipal Education Commission. We thank UniDT (Shanghai) Co., Ltd for the assistance in collecting data of Mandarin speakers.

	{\small
		\bibliographystyle{ieee_fullname}
		\bibliography{main}
	}


	\appendix
	\section{Statistics of the Dataset}
Due to the limitations of OpenPose \cite{cao2019openpose} in highly occluded cases, the pseudo labels for poses are noisy, especially for the hands and the mouth. Also, there are cases when multiple people appear in a frame, making it hard for the model to learn stable gestures. Therefore, we filter out a frame for any of the following cases: 
\begin{itemize}
    \item Some pseudo-label keypoints are significantly abnormal (e.g., lying on the top-left corner of the frame);
    \item Multiple humans are detected in a frame.
\end{itemize}
After the above data cleaning procedure, the statistics of the data we use are shown in Table \ref{tab:data_stat}.

\begin{table}[h]
    \centering
   	\small
    \caption{The statistics of the data we use.}
    \label{tab:data_stat}
    \begin{tabular}{lcccc}
        \toprule
        speaker & length (hours) & \#clips (train) & \#clips (val) \\
        \midrule
        Oliver & 12.02 & 119219 & 10546 \\
        Kubinec & 3.15 & 30100 & 3874 \\
        Luo & 7.46 & 73347 & 7254 \\
        Xing & 2.52 & 24757 & 2414 \\
        \bottomrule
    \end{tabular}
\end{table}

Among the four speakers, Oliver has the longest videos. Therefore, it is reasonable that the predictions for Oliver look the most vivid and natural. However, our results on Kubinec and Xing show that it is enough to learn the gesture style of a person with about two hours of videos.

\section{Additional Quantitative Results}

\begin{table}[h]
	\centering
	\small
	\caption{Quantitative results of our method with templates encoded by a VAE \cite{kingma2014vae} on seven other speakers of the Speech2Gesture \cite{ginosar2019learning} dataset.}
	\label{tab:additional_results}
	\begin{tabular}{lcccc}
		\toprule
		speaker & $L_2$ dist. & $\mathcal{E}_\text{lip}$ & FTD \\
		\midrule
		Almaram & 179.0 & 0.187 & 2.02 \\
		Angelica & 70.6 & 0.186 & 5.55 \\
		Conan & 124.6 & 0.213 & 0.626  \\
		Ellen & 110.4 & 0.239 & 1.56 \\
		Stewart & 79.0 & 0.347 & 0.893 \\
		Meyers & 64.0 & 0.195 & 1.88 \\
		Kagan & 144.4 & 0.133 & 1.38 \\
		\bottomrule
	\end{tabular}
\end{table}

Although the Speech2Gesture \cite{ginosar2019learning} dataset includes ten speakers (nine available online), not all of them suit our needs. For example, 
Almaram's hands are usually occluded so that the keypoints for hands and arms are out of view;
Angelica's videos may contain speech of other people on the phone, which brings ambiguities;
Kagan often walks around, which leads to noisy keypoints of the face and hands.
Nevertheless, we still train our model on each of them and show quantitative results in Table \ref{tab:additional_results}.

\end{document}